\documentclass[letterpaper,10pt,conference]{ieeeconf}

\IEEEoverridecommandlockouts
\usepackage{graphicx}
\usepackage{amsmath,amssymb}
\usepackage{booktabs}
\usepackage{multirow}
\usepackage{balance}
\usepackage{placeins}
\usepackage{capt-of}
\usepackage{microtype}
\usepackage{url}
\usepackage{xcolor}
\usepackage{tikz}
\usetikzlibrary{arrows.meta,positioning,fit}

\title{Bind-WAM: Binding-Uncertainty-Guided Selective Future Imagination\\
in World Action Models}

\author{Chenhao Qiu$^{1,4}$, Shuyi Chen$^{3}$, Songen Gu$^{1}$, Shufeng Nan$^{1}$, Tianyu Wang$^{1,2,4}$,\\
Siyu Lin$^{1,4}$, Yifan Li$^{1,2}$, Bingwen Zhu$^{1,2}$, Yanwei Fu$^{1,2\ddagger}$\thanks{$^{\ddagger}$Corresponding author.}\\[0.5em]
\normalsize $^{1}$Fudan University\quad $^{2}$Shanghai Innovation Institute\\
\normalsize $^{3}$Tongji University\quad $^{4}$ROBOTOPIA}

\begin{document}

\maketitle
\begin{tikzpicture}[remember picture,overlay]
  \node[anchor=north west,inner sep=0pt] at ([xshift=0.75in,yshift=-0.25in]current page.north west)
    {\includegraphics[width=1.15in]{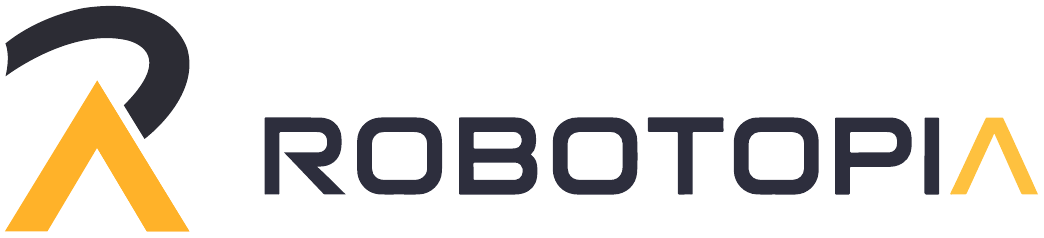}};
\end{tikzpicture}
\suppressfloats[t]
\thispagestyle{empty}
\pagestyle{empty}

\begin{abstract}
World Action Models (WAMs) can improve robot learning by jointly predicting future 
visual states and actions, but generating a visual future at every replanning
 step is expensive and sometimes unnecessary. This raises a critical question: \emph{when should a robot
policy imagine the future?} In this paper, we propose Bind-WAM, a binding-uncertainty-guided 
WAM for selective future imagination. Bind-WAM introduces language-object binding
uncertainty, selectively invoking joint future-action generation when the
current observation does not support stable task-relevant action generation. An offline 
 labeling pipeline uses complete demonstrations and action context to identify task-relevant objects and detect reference 
 mismatch, invisibility, and identity drift. The resulting pseudo-labels 
 supervise a lightweight router that predicts binding uncertainty from the 
 current observation and language instruction. At inference, a 
 shared WAM policy invokes selective future imagination according 
 to the predicted uncertainty and its change 
 between consecutive replanning steps. 
Simulation and real-world experiments demonstrate that Bind-WAM can
establish binding uncertainty as a strong, interpretable signal for
selective future imagination.
\end{abstract}

\section{Introduction}

\begin{figure}[!t]
    \centering
    \includegraphics[width=\columnwidth]{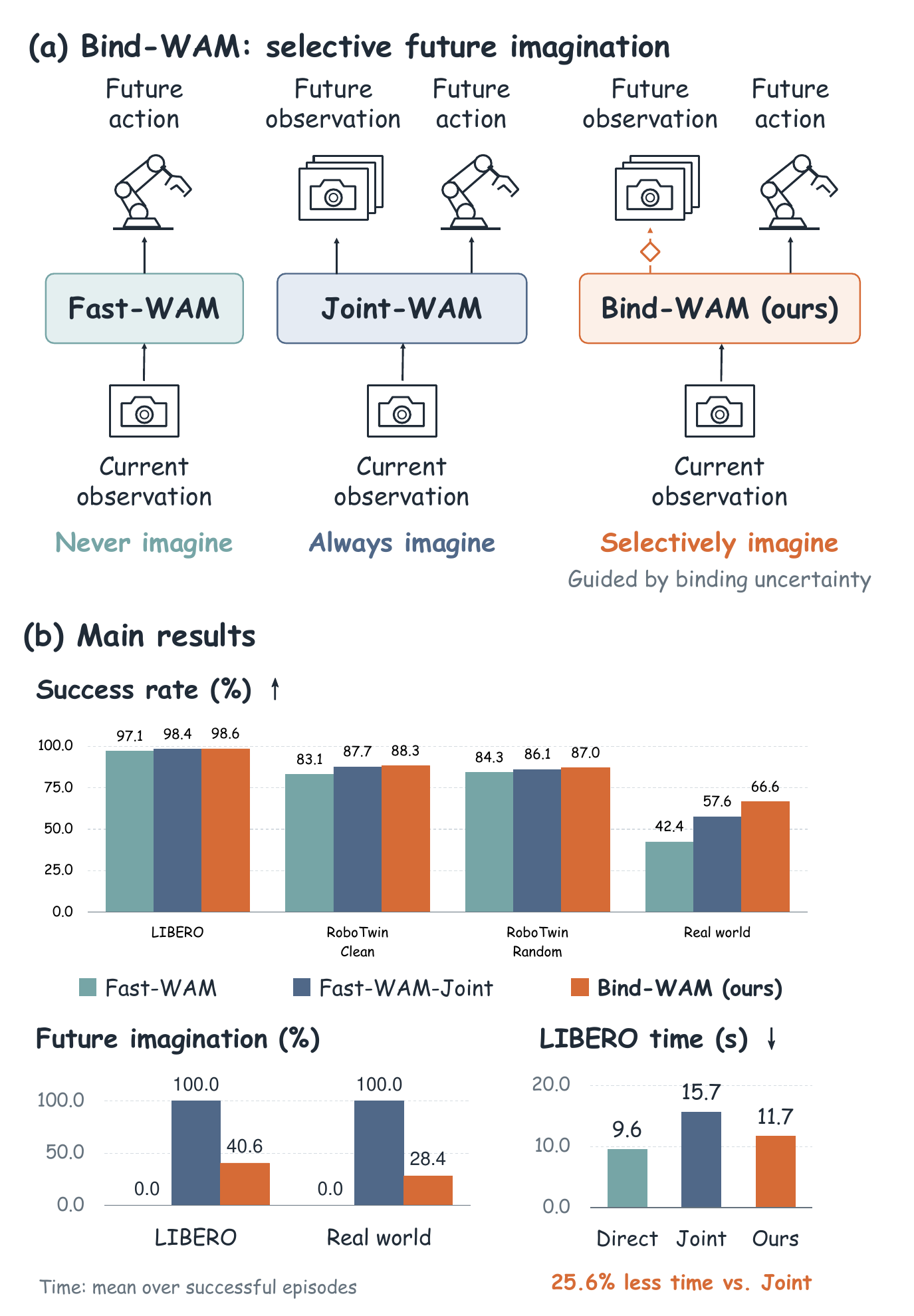}
    \caption{Illustration and main results of Bind-WAM. (a) Fast-WAM predicts
    actions without future generation, whereas Joint-WAM
    predicts future visual observations and actions. Bind-WAM uses binding
    uncertainty to selectively invoke joint future-action prediction.
    (b) Success rates across simulation and real-world tasks, future-imagination
    ratios, and LIBERO completion times. Bind-WAM achieves generally higher
    success rates than Fast-WAM and Fast-WAM-Joint, with 
    a minority of future imagination and much less completion time than
    Fast-WAM-Joint.}
    \label{fig:teaser}
\end{figure}

\begin{figure*}[t]
    \centering
    \includegraphics[width=\textwidth]{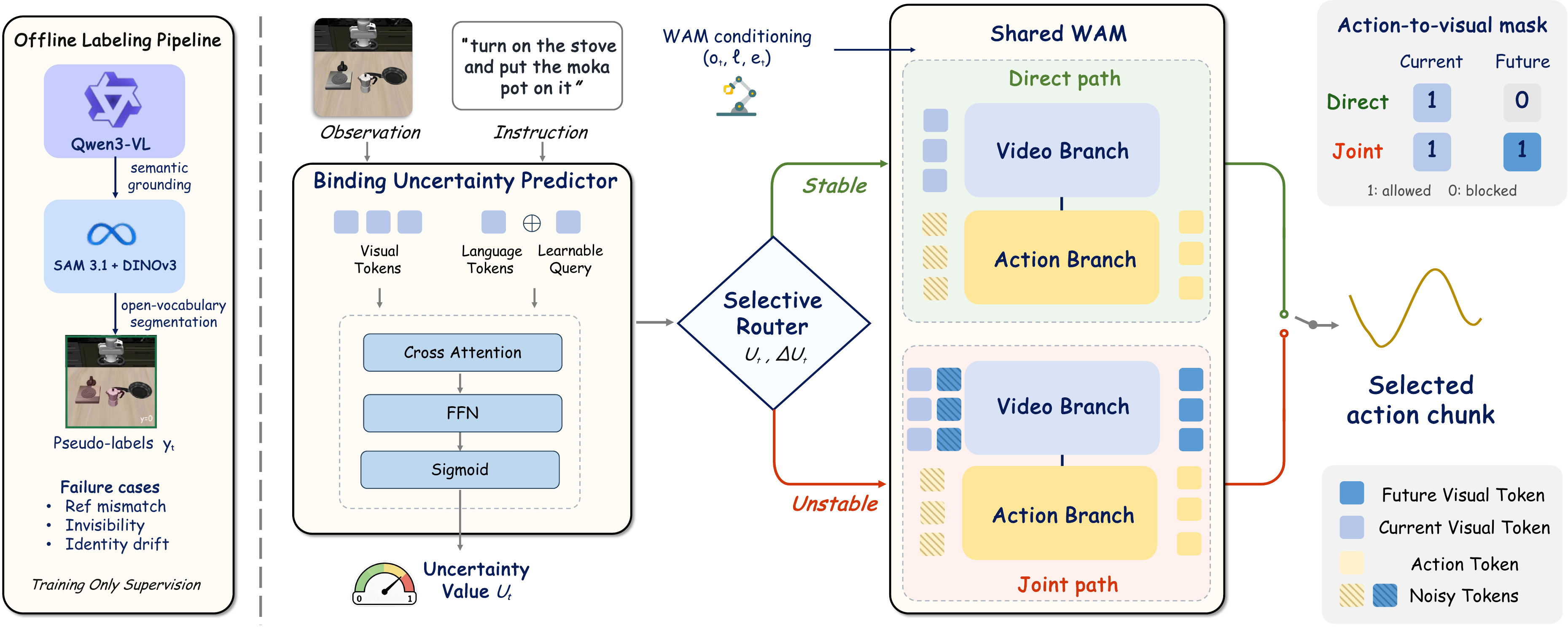}
    \caption{Overview of Bind-WAM. Offline binding labels capture reference
    mismatch, invisibility, and identity drift and supervise a lightweight
    predictor. At inference, the predictor estimates binding uncertainty $U_t$
    from the current observation and instruction. The router selects direct
    action prediction when uncertainty is low and stable, or joint future--action
    denoising when $U_t$ or its change $\Delta U_t$ exceeds a threshold.
    The inset summarizes action-to-visual attention:
    direct attends only to current visual tokens, whereas joint also attends
    to future visual tokens. Only the selected path
    executes and produces the action chunk. The offline labeling models are
    absent from online inference.}
    \label{fig:overview}
\end{figure*}

World Action Models (WAMs) improve robot policies with the ability to imagine
how a scene may evolve. By jointly predicting future visual states and actions,
they can exploit temporal and physical structure that is absent from a purely
reactive observation-to-action interface~\cite{ye2026dreamzero}. Yet traditional WAMs
always make an implicit assumption: future imagination should be triggered at every
replanning step. This uniform strategy ignores a basic property of
policy execution: the value of dreaming ahead is state dependent. In many states the
current observation already specifies the relevant objects and action, whereas
in others the policy must resolve missing, ambiguous, or unstable visual
evidence before acting.

This leads to the central question of this paper: \emph{when should
a robot policy imagine the future?} Existing work like Fast-WAM~\cite{yuan2026fastwam} shows that video co-training can benefit action prediction
even when future imagination is totally removed at deployment,
and Faster-WAM~\cite{zhao2026fasterwam} finds that future representations can improve robustness under
visual distribution shift. The important distinction
is therefore not whether future prediction is globally useful, but whether its
additional information is useful for the \emph{current} state. Recent methods
such as AdaWAM~\cite{tang2026adawam} enable adaptive inference by routing among
action-only, visual, and textual reasoning according to the execution context.
However, this identifies the need for future prediction mainly through execution
phases, but when imagination is truly necessary for the current state
remains unclear. Selective future imagination requires an interpretable signal to determine whether joint future-action prediction should be invoked.

We choose \emph{language-object binding uncertainty} as such a signal. A
language instruction assigns semantic roles to objects in the scene. For 
example, in
``place the bowl on the plate,'' the bowl is the affected object and the
plate is the relational reference. The policy's decision is fragile when a
required object is invisible, when similar instances compete for the same role,
or when an object's tracked identity changes during execution. These cases
are jointly defined by language and vision, arise before low-level action
generation, and directly threaten task completion, therefore
requiring future imagination to provide additional guidance. Stable language-object
bindings, in contrast, provide a clear setting for policies to skip future 
generation and predict correct actions.

In this paper, we propose Bind-WAM, a binding-uncertainty-guided WAM for selective future
imagination. Bind-WAM introduces an offline labeling pipeline that leverages
Qwen3-VL~\cite{bai2025qwen3vl}, SAM~3.1~\cite{carion2026sam3} and
DINOv3~\cite{simeoni2025dinov3} to generate uncertainty
labels representing role mismatch, invisibility, and identity drift, which are
used for training a 6.65M-parameter lightweight predictor. At deployment, the
predicted uncertainty value switches
one shared WAM between direct action prediction and joint future-action
denoising.


Our contributions are threefold. First, we formulate selective future 
imagination in World Action Models as a state-dependent
problem, and introduce language-object binding uncertainty as an
interpretable signal for deciding when joint future-action prediction should
be invoked. Second, we develop Bind-WAM, which constructs binding uncertainty
supervision from an offline labeling pipeline, and predicts uncertainty with 
a lightweight predictor to route a shared WAM between direct and joint 
inference. Third, simulation and real-world experiments demonstrate
that binding uncertainty can identify the states where future 
imagination is useful.

\section{Related Work}

\begin{figure*}[t]
    \centering
    \includegraphics[width=\textwidth]{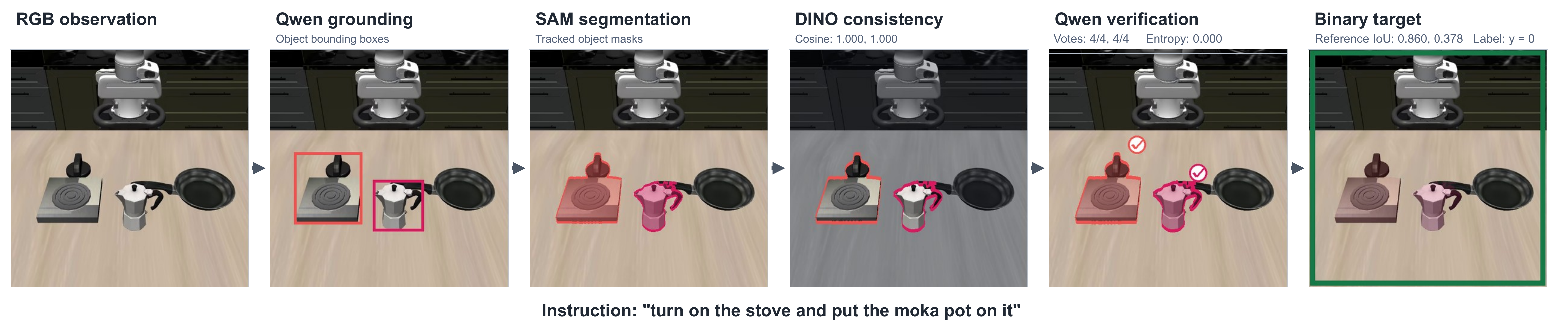}
    \caption{Offline binding-label construction pipeline
    for a typical LIBERO front-view image. From left to right, the
    RGB observation is passed to
    Qwen3-VL-32B grounding, which assigns role-specific boxes to the stove
    (red) and moka pot (magenta). SAM~3.1 produces tracked instance masks,
    DINOv3 measures their cosine consistency with reference features, and Qwen
    verification aggregates votes across temporal context. In this example,
    both roles are visible, their reference IoUs are $0.860/0.378$, their DINO
    similarities are $1.000/1.000$, and Qwen verification gives $4/4$ votes
    with zero entropy. These values yield the certain frame target $y_t=0$.}
    \label{fig:binding-annotation-pipeline}
\end{figure*}

\subsection{Future Imagination in World--Action Models}

World Action Models (WAMs) integrate visual world modeling with action
generation, allowing robot policies to exploit representations of how a scene
may evolve. A common design retains future prediction at inference, either
deriving actions from predictive future representations or jointly generating
futures and actions~\cite{ye2026dreamzero,hu2024vpp,kim2026cosmos,zhou2026tau0wm}.

Other methods remove explicit future generation or replace dense video
prediction with more compact inference mechanisms.
Fast-WAM~\cite{yuan2026fastwam} retains video co-training but skips future
prediction at deployment, showing that video modeling can benefit action
generation through training alone. ImageWAM~\cite{zhang2026imagewam} replaces
video generation with an image-editing prior and conditions its action expert
on editing-derived key-value caches, without decoding a target image at
inference. Faster-WAM~\cite{zhao2026fasterwam} preserves future-conditioned
representations through a single video-expert pass, then reuses them through
sparse video-action interaction during action denoising. These approaches
offer different choices about whether and how to retain future information.


Between always invoking future imagination and removing it entirely,
adaptive inference allows a WAM to decide when to imagine.
AdaWAM~\cite{tang2026adawam} uses execution-phase supervision to
selectively activate visual future reasoning and textual subtask reasoning
alongside action prediction.
RISE~\cite{lu2026rise} adapts latent rollout depth for autonomous driving
by balancing expected planning gains against computation cost, including
zero rollout when the observed context suffices.
Bind-WAM uses language-object binding uncertainty to route a shared WAM
between direct action prediction and joint future-action generation.
Its routing signal assesses whether task-relevant object roles are reliably
grounded in the current observation, tying selective imagination to
reference mismatch, invisibility, and identity drift.

\subsection{Language-Object Binding and Object-Centric Perception}

Language-object binding associates the entities and relational roles named
in an instruction with physical objects in the scene. Object-centric
semantic-spatial reasoning supports this association by selecting
task-relevant objects before predicting manipulation
affordances~\cite{luo2023grounding}. For world modeling,
SlotFormer~\cite{wu2023slotformer} learns temporal interactions among object
slots and predicts their future states, providing a structured alternative
to holistic scene representations.
Object-Centric World Model~\cite{jeong2025ocwm} extends object-centric prediction to
language-guided manipulation by forecasting future slot representations
conditioned on the instruction. 
OA-WAM~\cite{liu2026oawam} instead makes object
identity explicitly addressable by separating each slot into a persistent
identity address and time-varying content. 

Bind-WAM uses object-centric information for uncertainty supervision rather
than as the state space of its world model. We construct labels from the
consistency of instruction-role assignments to observed objects.
These labels train a predictor to recognize unstable bindings from the
current observation and instruction, without requiring object-slot
extraction in the control loop.
Bind-WAM uses this uncertainty supervision to decide
when future imagination should be invoked.

\section{Method}

Figure~\ref{fig:overview} summarizes the overall
architecture of Bind-WAM, showing how our approach connects offline binding
supervision to online selective future imagination. The uncertainty predictor
evaluates the current observation and instruction before WAM execution,
and its output determines which inference path the shared policy uses.

\subsection{Problem Formulation}
Given a current visual observation $o_t$, proprioceptive state $e_t$,
and language instruction $\ell$, we aim to learn a policy that predicts
an executable action chunk $a_{t:t+H}$:
\begin{equation}
\pi(a_{t:t+H}\mid o_t,e_t,\ell).
\end{equation}

\subsection{Offline Uncertainty Labeling Pipeline}

Manual frame-level annotation of language roles and visual entities is costly, so we construct pseudo-labels from demonstrations in an automated pipeline. 
Given a demonstration trajectory and its language instruction, we first identify 
task-relevant entities, including the manipulated object and relational
anchors using Qwen3-VL-32B. To determine which candidate corresponds to each 
task role, Qwen examines mask overlays from multiple temporal context frames, including 
interaction-relevant frames.
SAM~3.1 then generates candidate instance masks from entity-specific text prompts
and propagates them across the trajectory. DINOv3
ViT-L/16 features are then pooled inside each mask to measure identity consistency.
The complete annotation flow is illustrated in Fig.~\ref{fig:binding-annotation-pipeline}.

We uniformly sample 32 frames from each complete demonstration for labeling. For role $r$, let
$\mathrm{IoU}_r$ be the overlap between the Qwen-verified reference mask and
the original candidate track at the reference frame, $v_t^r$ the current
visibility, and $s_t^r$ the cosine similarity between current and reference
DINO features. The role-level uncertainty label is defined as
\begin{equation}
y_t^r=
\begin{cases}
1, & \mathrm{IoU}_r<0.3\;\lor\;v_t^r=0\;\lor\;s_t^r<0.9,\\
0, & \text{otherwise},
\end{cases}
\end{equation}
which means a role is labeled uncertain at frame $t$ if its reference mask
is unreliable, the tracked object is not visible, or its DINO feature similarity to 
the verified reference is relatively low.
The frame-level label takes the most uncertain task-relevant role:
\begin{equation}
y_t=\max_r y_t^r.
\end{equation}
This label captures three concrete binding failures: reference mismatch,
invisibility, and identity drift.

These binary labels supervise the uncertainty predictor using only the current RGB observation and language instruction. Qwen, SAM, DINO, action signals, and
trajectory context are used exclusively during offline annotation and are not required at inference time.

\subsection{Binding Uncertainty Predictor}

The predictor is trained on offline labeled data to predict frame-level binding 
uncertainty $U_t$ from the current observation $o_t$ and language instruction $\ell$:
\begin{equation}
U_t=f_\theta(o_t,\ell), \qquad U_t\in[0,1],
\label{eq:uncertainty}
\end{equation}
where high $U_t$ denotes unstable binding between instruction roles and objects in the scene. 


The predictor follows a lightweight ViT-like architecture~\cite{dosovitskiy2021vit}, and training loss
combines class-weighted binary cross entropy with a Brier term~\cite{brier1950verification}:
\begin{equation}
\mathcal{L}_{\mathrm{bind}}=\mathrm{BCE}_{w}(\hat U_t,y_t)+\lambda(\hat U_t-y_t)^2.
\end{equation}

\begin{figure*}[t]
\centering
\includegraphics[width=\textwidth]{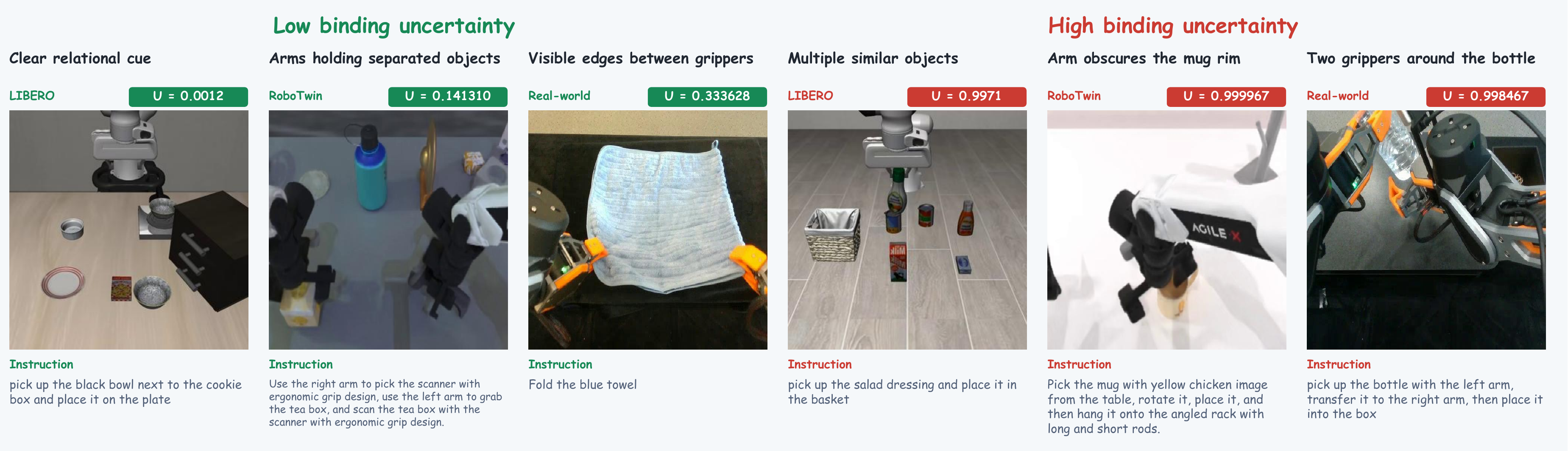}
\caption{Representative binding-uncertainty predictions. The low-uncertainty
examples show a clear relational cue, arms holding separated objects,
and towel edges visible between grippers. The high-uncertainty examples show multiple
similar objects, an arm obscuring the mug rim, and two grippers around
a bottle.
Each panel reports the task instruction and predicted binding uncertainty.}
\label{fig:examples}
\end{figure*}

\subsection{Binding-Uncertainty-Guided WAM Routing}

Bind-WAM realizes two inference paths. The
\textbf{direct path} $\pi_{\mathrm{direct}}$ uses first-frame-only action-to-video 
attention, where action tokens can
attend to the observed frame but not to denoised future-video tokens. The \textbf{joint path} $\pi_{\mathrm{joint}}$ uses full action-to-video
attention, which jointly denoises future visual latents and the action chunk.

We implement both paths within a single parameter-sharing 
WAM rather than maintaining separate policies for the two paths.
Both paths share the video expert, action expert, and mixed transformer, and 
differ only in their action-to-video attention masks. Let the video token 
sequence be
$Z^v=[Z_0^v,\ldots,Z_H^v]$, where $Z_0^v$ represents the current observation 
and $Z_{1:H}^v$ represent future states. Let $Z^a$ denote the action tokens. The
direct mask allows the action tokens to attend only to the current frame:

\begin{equation}
M_{\mathrm{direct}}(Z^a,Z_h^v)=\begin{cases}
1, & h=0,\\
0, & h>0.
\end{cases}
\end{equation}

In contrast, the joint mask allows the action token to attend to all video 
tokens:

\begin{equation}
M_{\mathrm{joint}}(Z^a,Z_h^v)=1,\qquad h=0,\ldots,H.
\end{equation}

Consequently, the direct path denoises actions directly from the current 
observation, whereas the joint path can exploit future visual representations 
through future latent denoising.

\subsection{Selective Future Imagination}

At time $t$ during inference, the policy receives observation
$o_t$, robot state $e_t$,
and language instruction $\ell$. The policy can either predict actions directly or
jointly denoise future visual and action latents. Bind-WAM treats this choice as a
state-dependent routing problem and uses
binding uncertainty $U_t$ as a semantic signal that can be evaluated from current
observation and language instruction. Instead of only considering absolute values, we 
additionally track the uncertainty change between consecutive timesteps:
\begin{equation}
\Delta U_t=\lvert U_t-U_{t-1}\rvert.
\end{equation}

The shared policy follows selective future imagination by choosing different paths:
\begin{equation}
\pi(a_{t:t+H}\mid o_t,e_t,\ell)=
\begin{cases}
\pi_{\mathrm{joint}}, & U_t\geq\tau_U\;\lor\;\Delta U_t\geq\tau_\Delta, \\
\pi_{\mathrm{direct}}, & \text{otherwise}.
\end{cases}
\end{equation}


\subsection{Training Objective}
We train Bind-WAM with flow-matching objectives~\cite{lipman2023flowmatching} for normalized action
  chunks $y^a$ and future-video latents $y^v$.
  For each modality $d\in\{a,v\}$, we independently sample Gaussian noise
  $\epsilon^d\sim\mathcal{N}(0,I)$ and a noise level
  $t_d\sim p_{\mathrm{shift}}$, then construct
  \begin{equation}
  y^d_{t_d}=(1-t_d)y^d+t_d\epsilon^d,
  \qquad v^d=\epsilon^d-y^d.
  \end{equation}
  The shared WAM predicts both velocity fields under the
  binding-dependent attention mask $M_g$:
  \begin{equation}
  (\hat v^a,\hat v^v)
  =
  F_\phi(y^a_{t_a},y^v_{t_v},t_a,t_v,c;M_g),
  \end{equation}
  where $c$ contains the current visual observation, proprioceptive state,
  and language instruction. The training objective is
  \begin{equation}
  \mathcal{L}_{\mathrm{WAM}}
  =\lambda_a\mathcal{L}_a+\lambda_v\mathcal{L}_v,
  \end{equation}
  where each modality-specific loss is
  \begin{equation}
  \mathcal{L}_d
  =\mathbb{E}\!\left[
  w(t_d)\operatorname{MSE}
  (\hat v^d,\epsilon^d-y^d)
  \right].
  \end{equation}
  Here, $p_{\mathrm{shift}}$ and $w$
  are the timestep sampling distribution and loss weighting inherited
  from the WAM scheduler.

\section{Experiments}

\begin{table}[!ht]
\centering
\caption{Binding uncertainty predictor performance on episode-disjoint test
sets from LIBERO and RoboTwin.}
\label{tab:router}
\small
\setlength{\tabcolsep}{7pt}
\renewcommand{\arraystretch}{1.10}
\begin{tabular}{lrr}
\toprule
Metric & LIBERO & RoboTwin \\
\midrule
Balanced Accuracy & 87.59\% & 96.75\% \\
AUROC & 0.9653 & 0.9984 \\
AUPRC & 0.9917 & 0.9999 \\
Brier score & 0.0504 & 0.0070 \\
Uncertain recall @ 0.5 & 97.32\% & 99.27\% \\
Certain recall @ 0.5 & 77.86\% & 94.23\% \\
\bottomrule
\end{tabular}
\end{table}

\begin{figure*}[t]
\begin{minipage}[t]{\columnwidth}
\vspace{0pt}
\centering
    \includegraphics[width=\columnwidth]{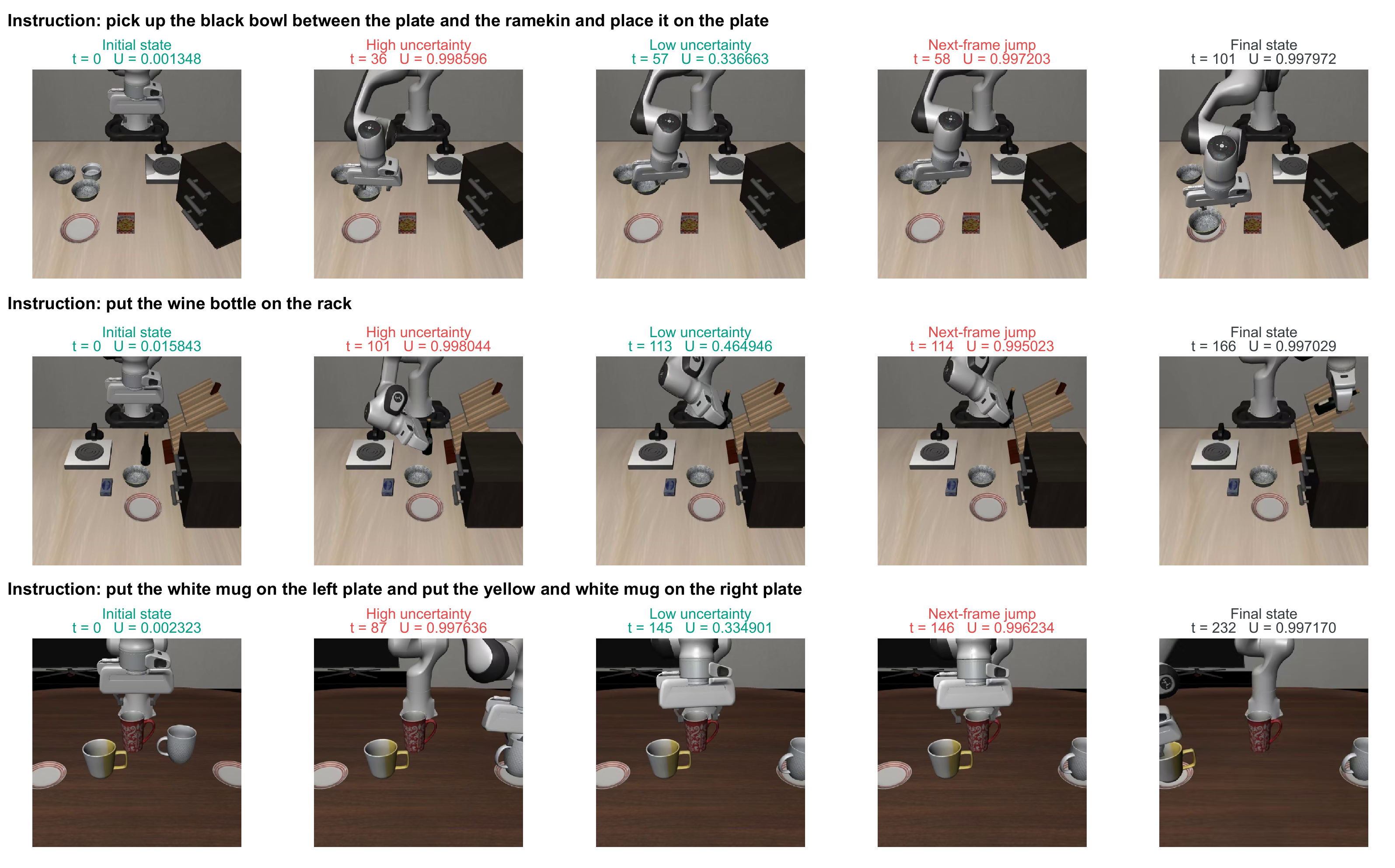}
    \caption{Binding uncertainty varies throughout task execution. Each row
    shows one selected LIBERO episode, with frame indices $t$
    and uncertainty values $U$. Unstable states include high uncertainties
    or jumps between frames.}
    \label{fig:uncertainty-evolution}
\end{minipage}\hfill
\begin{minipage}[t]{\columnwidth}
\vspace{0pt}
\centering
    \includegraphics[width=\columnwidth]{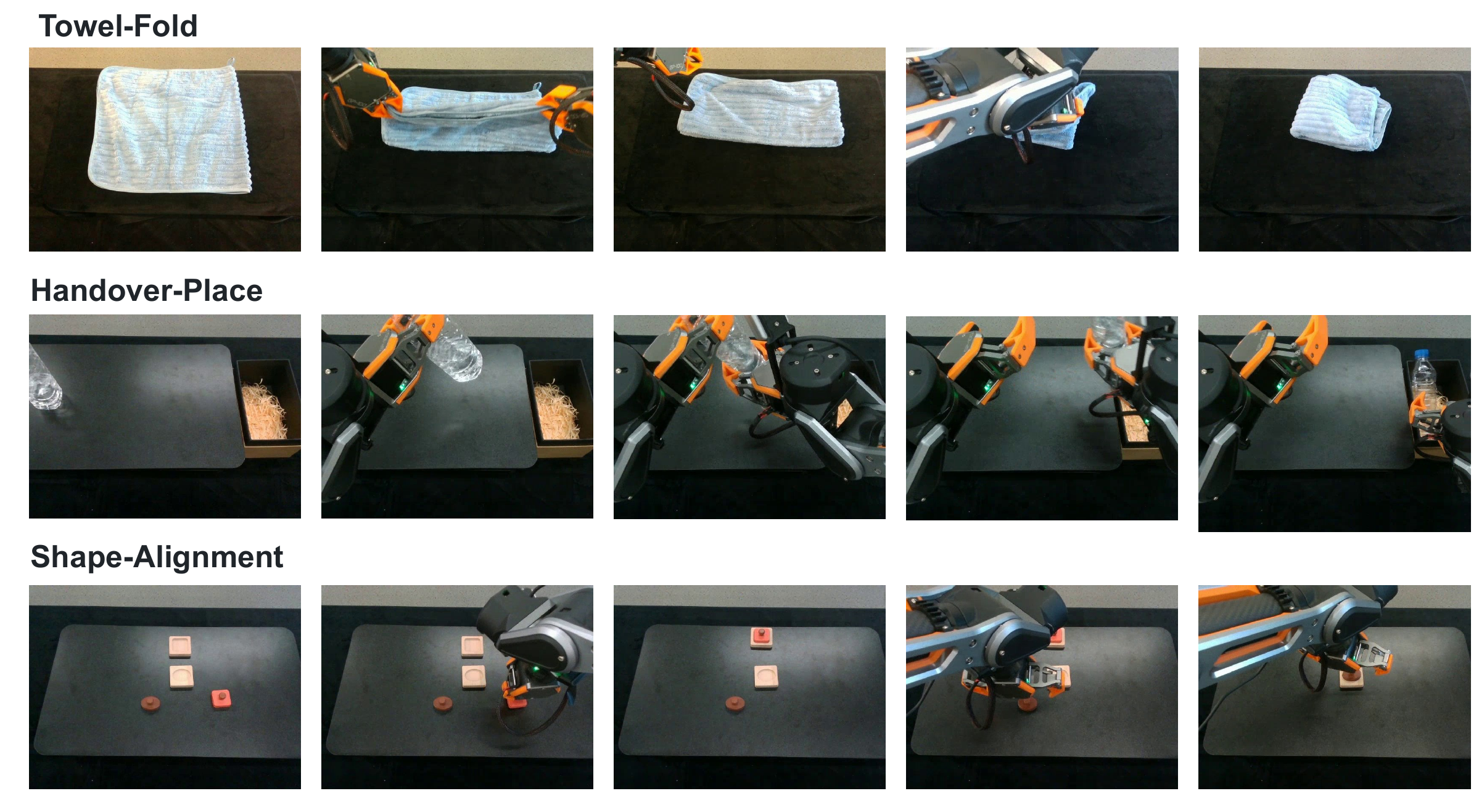}
    \caption{Representative successful executions of the three real-world
    tasks. Each row shows five selected
    frames in temporal order from left to right. Each frame shows the overhead camera view. Frames are selected by
    manipulation stage rather than at uniform time intervals.}
    \label{fig:realworld-demonstrations}
\end{minipage}
\par\vspace{10pt}
\centering
\captionof{table}{LIBERO evaluation results. Success rates are percentages and time
denotes completion time averaged over successful episodes. $^\dagger$ denotes results
reproduced under our matched setup.}
\label{tab:libero}
\small
\setlength{\tabcolsep}{9pt}
\renewcommand{\arraystretch}{1.08}
\begin{tabular}{lrrrrrrr}
\toprule
Method & Spatial & Object & Goal & Long & Mean & Time (s) & Future Imagination (\%) \\
\midrule
OpenVLA & 84.7 & 88.4 & 79.2 & 53.7 & 76.5 & -- & -- \\
$\pi_0$ & 96.8 & 98.8 & 95.8 & 85.2 & 94.2 & -- & -- \\
$\pi_{0.5}$ & \underline{98.8} & 98.2 & 98.0 & 92.4 & 96.9 & -- & -- \\
LingBot-VA & 98.5 & \underline{99.6} & 97.2 & \underline{98.5} & \underline{98.5} & -- & -- \\
Motus & 96.8 & \textbf{99.8} & 96.6 & 97.6 & 97.7 & -- & -- \\
AdaWAM & 98.0 & \underline{99.6} & 97.1 & \textbf{99.1} & \underline{98.5} & -- & -- \\
\midrule
Fast-WAM$^\dagger$ & 97.6 & 99.4 & 96.6 & 94.8 & 97.1 & 9.6 & 0.0 \\
Fast-WAM-Joint$^\dagger$ & \textbf{99.8} & 98.8 & \underline{98.6} & 96.4 & 98.4 & 15.7 & 100.0 \\
Bind-WAM(Ours)  & \textbf{99.8} & 99.2 & \textbf{98.8} & 96.6 & \textbf{98.6} & 11.7 & 40.6 \\
\bottomrule
\end{tabular}
\end{figure*}

\begin{figure*}[t]
\centering
\captionof{table}{RoboTwin~2.0 success rates (\%). Success rates are percentages and time
denotes completion time averaged over successful episodes. $^\dagger$ denotes results
reproduced under our matched setup.}
\label{tab:robotwin}
\begingroup
\fontsize{8.5}{10}\selectfont
\renewcommand{\arraystretch}{1.02}
\setlength{\tabcolsep}{3pt}
\begin{tabular}{l*{10}{r}}
\toprule
Task & GO-1 & $\pi_0$ & $\pi_{0.5}$ & X-VLA & LingBot-VA & Motus & AdaWAM & Fast-WAM$^\dagger$ & Fast-WAM-Joint$^\dagger$ & Bind-WAM \\
\midrule
\multicolumn{11}{c}{Clean} \\
\midrule
Hanging Mug & 0.0 & 14.0 & 18.0 & 23.0 & 40.0 & 38.0 & 59.0 & 57.0 & \textbf{72.0} & \underline{71.0} \\
Pick Diverse Bottles & 61.0 & 69.0 & 81.0 & 58.0 & \underline{89.0} & \textbf{90.0} & 87.0 & 82.0 & 86.0 & 87.0 \\
Put Object Cabinet & 60.0 & 85.0 & 80.0 & 46.0 & 80.0 & 88.0 & \textbf{96.0} & 94.0 & \underline{95.0} & \underline{95.0} \\
Rotate QR Code & 22.0 & 74.0 & 89.0 & 34.0 & \textbf{96.0} & 89.0 & \underline{94.0} & 93.0 & 91.0 & 92.0 \\
Scan Object & 1.0 & 55.0 & 72.0 & 14.0 & \textbf{96.0} & 67.0 & \underline{92.0} & 89.0 & \underline{92.0} & 89.0 \\
Stack Bowls Three & 4.0 & 77.0 & 77.0 & 76.0 & \textbf{100.0} & 79.0 & \textbf{100.0} & 77.0 & 82.0 & \underline{87.0} \\
Stamp Seal & 19.0 & 46.0 & 79.0 & 76.0 & \underline{96.0} & 93.0 & 91.0 & 90.0 & \underline{96.0} & \textbf{97.0} \\
\midrule
Average SR & 23.9 & 60.0 & 70.9 & 46.7 & 85.3 & 77.7 & \textbf{88.4} & 83.1 & 87.7 & \underline{88.3} \\
Average Time & -- & -- & -- & -- & -- & -- & -- & 23.2 & 25.9 & 24.4 \\
Future Imagination (\%) & -- & -- & -- & -- & -- & -- & -- & 0.0 & 100.0 & 44.5 \\
\midrule
\multicolumn{11}{c}{Random} \\
\midrule
Hanging Mug & 0.0 & 11.0 & 17.0 & 27.0 & 28.0 & 38.0 & \underline{60.0} & \textbf{62.0} & 58.0 & \underline{60.0} \\
Pick Diverse Bottles & 56.0 & 31.0 & 71.0 & 36.0 & 82.0 & \textbf{91.0} & 86.0 & 85.0 & 87.0 & \underline{88.0} \\
Put Object Cabinet & 43.0 & 87.0 & 79.0 & 48.0 & 79.0 & 71.0 & \textbf{91.0} & 87.0 & \textbf{91.0} & \underline{89.0} \\
Rotate QR Code & 9.0 & 70.0 & 87.0 & 33.0 & 91.0 & 73.0 & \textbf{94.0} & 88.0 & 90.0 & \underline{93.0} \\
Scan Object & 2.0 & 42.0 & 65.0 & 36.0 & \underline{91.0} & 66.0 & \underline{91.0} & 90.0 & \textbf{93.0} & \textbf{93.0} \\
Stack Bowls Three & 7.0 & 75.0 & 71.0 & 86.0 & \textbf{98.0} & 87.0 & \textbf{98.0} & 84.0 & \underline{88.0} & 87.0 \\
Stamp Seal & 13.0 & 33.0 & 55.0 & 82.0 & \underline{97.0} & 92.0 & 86.0 & 94.0 & 96.0 & \textbf{99.0} \\
\midrule
Average SR & 18.6 & 49.9 & 63.6 & 49.7 & 80.9 & 74.0 & \underline{86.6} & 84.3 & 86.1 & \textbf{87.0} \\
Average Time & -- & -- & -- & -- & -- & -- & -- & 24.0 & 26.4 & 25.3 \\
Future Imagination (\%) & -- & -- & -- & -- & -- & -- & -- & 0.0 & 100.0 & 36.2 \\
\bottomrule
\end{tabular}
\endgroup
\end{figure*}

\subsection{Experimental Setup}

\textbf{Implementation Details.}
For the offline labeling pipeline, we uniformly sample 32
frames from each valid episode and label them with $0/1$ uncertainty tags.
We only label the top-head, front or central camera views in all experiments
for efficiency. 
To train the uncertainty predictor, we split episodes, rather than individual
frames, into 70\%/15\%/15\% training, validation, and test partitions. This
prevents temporally adjacent observations
from crossing split boundaries.

The binding predictor receives a $224\!\times\!224$ RGB image and the task
instruction. The input image is divided into $16\times16$ patches, and instruction tokens 
are embedded by a lightweight learned vocabulary and mean-pooled. The language vector is then added to a learnable query token. Four image-query attention blocks with hidden
size 256 exchange information between patch tokens and the query. Image and
query streams use separate projections and feed-forward networks. A two-layer
MLP followed by a sigmoid maps the final query to binding uncertainty $U_t$. 
The lightweight predictor contains around 6.65M parameters. All uncertainty 
predictors are trained for 15 epochs with batch size 16, and we select the best checkpoint with the lowest validation BCE.

For the policy, binding uncertainty values at consecutive replanning states
are precomputed for the demonstrations. Direct and joint paths are selected
during training by the same routing rule used at deployment and update one
parameter-sharing WAM. We use pretrained Wan2.2-5B~\cite{wan2025} as the video backbone in the WAM.

Values of thresholds $\tau_U$ and
$\tau_\Delta$ decide the fraction of inference with joint future-action prediction. For 
example, in LIBERO, setting $\tau_U=0.9976$ and $\tau_\Delta=0.25$ selects 
40.58\% of replanning states for the joint path. The policy replans every 10 environment steps.
Qwen3-VL, SAM~3.1, and DINOv3 are used only
to construct offline labels and are absent from the control loop.

\textbf{Simulation Benchmarks.}
We evaluate Bind-WAM on LIBERO~\cite{liu2023libero} and RoboTwin~2.0~\cite{chen2025robotwin} for single-arm and
bimanual manipulation evaluation. The LIBERO training data
contains 1,712 episodes
from the Spatial, Object, Goal, and Long suites. Each suite 
comprises 10 tasks,
and we evaluate every task over 50 trials. For RoboTwin~2.0, we select
seven tasks following~\cite{tang2026adawam},
and collect 50 clean plus 500 randomized
demonstrations per task. We evaluate every
policy on 100 trials per task under both clean and randomized conditions.
All models are trained for 10 epochs on both benchmarks.

\textbf{Real-World Setup.}
Our real-world platform consists of a HEXFELLOW ARCHER-D6Y bimanual system,
one overhead Intel
RealSense D435 camera, and two wrist-mounted D405 cameras. The policy takes all
three views, dual-arm joint angles and gripper states as input and predicts
joint-angle action chunks for both arms. We evaluate on three bimanual manipulation tasks:
Fold a towel(Towel-Fold); Hand over a bottle and place it into a box(Handover-Place);
and Pick up the object with correct shape
to match the corresponding slot(Shape-Alignment). 
We collect 75 real demonstrations for each task, and evaluate over 11 trials. All models are trained for 10 epochs on the three tasks. For real-world deployment, we use a workstation with an NVIDIA RTX 6000 Ada GPU. 

\begin{figure}[t]
\centering
\includegraphics[width=\columnwidth]{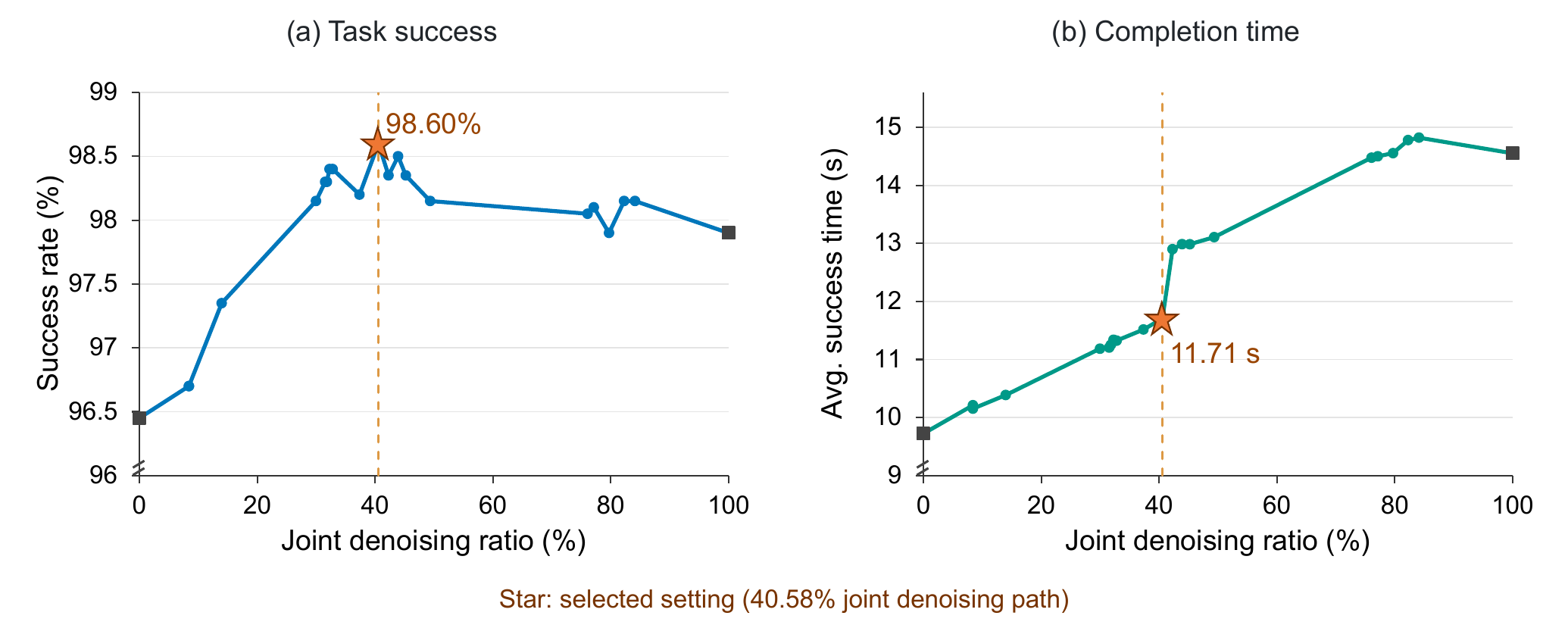}
\caption{Effect of joint denoising in Bind-WAM on LIBERO. Joint denoising ratio
    is the fraction of replanning steps executing the joint path. The panels
    show success rate and average completion time over successful episodes.
    Circles show the tested threshold
    settings, squares mark direct-only and joint-only execution, and stars highlight the selected setting ($\tau_U=0.9976$ and 
    $\tau_\Delta=0.25$). The direct-only and joint-only endpoints
    use the router-trained Bind-WAM checkpoint, whereas the
    Fast-WAM and Fast-WAM-Joint baselines in Table II are trained
    under their fixed inference modes.}
    \label{fig:libero-rollout-sweep}
\end{figure}

\begin{figure*}[t]
\begin{minipage}[t]{\columnwidth}
\vspace{0pt}

\centering
\captionof{table}{Real-world bimanual task success rates. Success rates are percentages and time
denotes completion time averaged over successful episodes.}
\small
\setlength{\tabcolsep}{2pt}
\begin{tabular}{lcccc}
\toprule
Task & Motus & FastWAM & FastWAM-Joint & Ours \\
\midrule
Towel-Fold & 54.5 & 45.5 & 63.6 & \textbf{72.7} \\
Handover-Place & 63.6 & 54.5 & \textbf{72.7} & \textbf{72.7} \\
Shape-Alignment & 36.4 & 27.3 & 36.4 & \textbf{54.5} \\
\midrule
Average SR &  51.5 & 42.4 & 57.6 & \textbf{66.6} \\
Average Time & 50.1  & 37.1 & 41.9 & 39.0 \\
Future Imagination (\%) & 100.0  & 0.0 & 100.0 & 28.4 \\
\bottomrule
\end{tabular}%
\label{tab:realworld_results}

\end{minipage}\hfill
\begin{minipage}[t]{\columnwidth}
\vspace{0pt}
\centering
  \captionof{table}{Routing ablation on LIBERO.
  Success rates are percentages and time
denotes completion time averaged over successful episodes.}
  \label{tab:routing_ablation}
  \small
  \setlength{\tabcolsep}{3pt}
  \renewcommand{\arraystretch}{1.10}
  \begin{tabular}{lrrr}
  \toprule
  Routing strategy & \shortstack{Success\\rate} & Time (s) & \shortstack{Future\\Imagination (\%)} \\
  \midrule
  Uncertainty router & \textbf{98.60} & 11.71 & 40.58 \\
  Random routing     & 97.35          & 14.58 & 40.89 \\
  Uniform routing    & 97.50          & 14.64 & 40.58 \\
  \bottomrule
  \end{tabular}%

\end{minipage}
\end{figure*}

\begin{figure}[t]
\vspace{0pt}
\centering
    \includegraphics[width=\columnwidth]{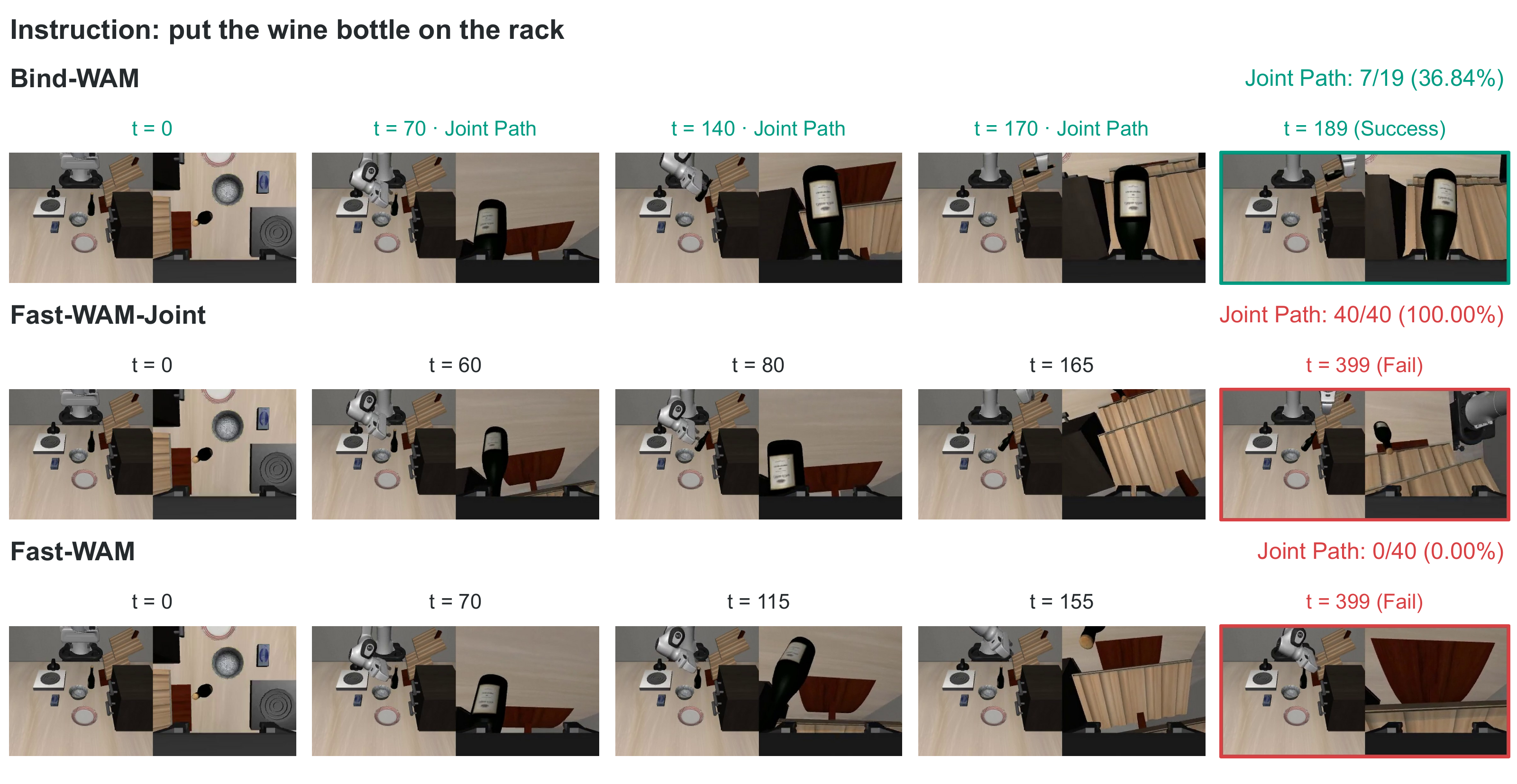}
    \caption{Comparison of different methods on the LIBERO wine-rack task.
    Each row shows five selected keyframes with policy steps labeled;
    ``Joint Path'' labels identify selected Bind-WAM joint-inference
    steps.}
    \label{fig:imagination-evidence}
\end{figure}

\subsection{Main Results}

\textbf{Binding Uncertainty Predictor Performance.}
Table~\ref{tab:router} reports uncertainty prediction performance under
LIBERO and RoboTwin~2.0 settings. The predictor performs strongly on both
benchmarks, consistently separating uncertain from certain bindings while
maintaining high recall for both regimes. Performance is particularly strong
on RoboTwin~2.0, where the lower Brier score also indicates more accurate
probability estimates. Overall,
the results show that the lightweight predictor design can capture binding
uncertainty across both single-arm and bimanual manipulation settings.


Figure~\ref{fig:examples} shows low and high predicted binding uncertainty
across LIBERO, RoboTwin, and real-world scenes. 
Observations with high uncertainty $U_t$ show ambiguous language-object
bindings, which may lead to incorrect actions if the policy skips future imagination, whereas observations with
low uncertainty provide clear visual evidence, indicating that direct action
prediction may already be reliable.
Figure~\ref{fig:uncertainty-evolution} illustrates how binding uncertainty
changes within an episode. In these examples,
high uncertainty and sharp changes occur around contact-rich, complex
manipulation stages, motivating future imagination to provide additional
guidance when the current visual evidence is unstable.

\textbf{Simulation Results.}
Table~\ref{tab:libero} summarizes the evaluation results on LIBERO.
Bind-WAM achieves the highest mean success rate of 98.60\% among the compared VLA or WAM methods~\cite{yuan2026fastwam,tang2026adawam,kim2024openvla,black2025pi0,intelligence2025pi05,li2026lingbotva,bi2025motus}, while invoking the
joint denoising path for future prediction at an average of 40.58\% for all replanning states. It improves
over reproduced Fast-WAM by 1.50 points and
also exceeds the reproduced Fast-WAM-Joint baseline by 0.20 points, meanwhile 
reducing average task completion time from 15.73~s for Fast-WAM-Joint to
11.71~s, a 25.6\% reduction. These results indicate that Bind-WAM can identify
the states where future imagination is genuinely useful for task execution.

Table~\ref{tab:robotwin} reports results on the seven RoboTwin~2.0 bimanual
manipulation tasks under clean and randomized conditions. Bind-WAM achieves
the second-highest average success rate of 88.29\% under clean conditions
and the highest of 87.00\% under randomized conditions among the compared
methods~\cite{yuan2026fastwam,tang2026adawam,black2025pi0,intelligence2025pi05,li2026lingbotva,bi2025motus,agibot2025go1,zheng2026xvla}.
Bind-WAM improves average success over Fast-WAM by 5.15 and 2.71 points.
It also exceeds Fast-WAM-Joint by 0.58 and 0.86 points while reducing average
completion time over successful episodes by more than 1~s on both settings, with future imagination triggered
at 44.45\% and 36.20\% of replans respectively. These results extend the
benefit of selective future imagination in Bind-WAM to bimanual manipulation.

\textbf{Real-World Results.}
Table~\ref{tab:realworld_results} reports success rates on
11 real-robot trials for all models on each task. Our 
Bind-WAM achieves
the highest average success rate of 66.6\%, showing benefit across deformable object manipulation,
bimanual handover, and precise alignment.
Bind-WAM invokes future imagination at 28.4\% of replanning steps,
while reducing average completion time over successful episodes from
41.9~s for Fast-WAM-Joint to 39.0~s.
These results support the effectiveness of binding-uncertainty-guided selective
future imagination in different real-world scenarios.
Figure~\ref{fig:realworld-demonstrations}
illustrates successful executions
of the three real-world tasks that highlight bimanual coordination and contact-rich manipulation.

\subsection{Analysis}
\textbf{Effect of Joint Denoising Ratio.}
Figure~\ref{fig:libero-rollout-sweep} examines how the joint video-action denoising ratio
affects the performance of Bind-WAM on LIBERO as the uncertainty threshold 
varies. Success rate improves
from 96.45\% with direct-path only execution to 98.15--98.40\% at approximately
30--33\% joint denoising, and reaches its highest observed value of 98.60\%
at 40.58\% joint denoising. Increasing the ratio further provides no additional success gain, despite generally longer completion times. This non-monotonic
trend supports selective future imagination: executing the joint path more
often does not necessarily produce a better policy.

Instead of the selected setting, a lower-cost setting
at 32.81\% joint denoising retains 98.40\% success with an average completion time
of 11.32~s, illustrating the flexibility of our selective imagination approach.

\textbf{Effect of Routing Strategy.}
Table~\ref{tab:routing_ablation} compares different
routing strategies at nearly identical ratios for future prediction
(40.58--40.89\%). Random routing selects the joint path randomly at each
replanning step
with a probability of 0.4058, while uniform routing selects the 
joint path at replanning positions spaced approximately 1/0.4058 steps apart (rounding each target position to the nearest integer).
The binding-uncertainty routing strategy in Bind-WAM achieves 98.60\% success, exceeding
random and uniform routing by 1.25 and 1.10 points respectively.
It also reduces average completion time over successful episodes from
14.58--14.64~s to 11.71~s, a reduction of approximately 20\%. These results support
binding uncertainty as an effective signal for selecting efficient paths to invoke
future prediction, avoiding redundant imagination that may not contribute to task execution.

\textbf{When Bind-WAM Invokes Imagination.}
Figure~\ref{fig:imagination-evidence} compares Bind-WAM with the
Fast-WAM and Fast-WAM-Joint baselines on the LIBERO task
``put the wine bottle on the rack,'' starting from the same initial scene.
Bind-WAM succeeds at $t=189$, invoking joint denoising at 7 of 19 replans
(36.84\%), while both baselines fail at $t=399$.
The selected Bind-WAM keyframes show joint inference at $t=70$,
$140$, and $170$ as the bottle is manipulated toward
the rack.

The failure of Fast-WAM illustrates the limitation of removing future
imagination in this episode. At $t=70$ when the arm is going
to grasp the wine bottle, the policy skips
future imagination and fails to predict the correct action,
so the arm misses the bottle.
The failure of Fast-WAM-Joint further shows
that invoking future imagination at every step is also insufficient for success. At $t=60$ when the arm is approaching the bottle, the
joint future-action generation predicts a biased action and knocks
the bottle over. In contrast,
Bind-WAM succeeds by triggering the joint path at $t=70$, which supports the value
of identifying states where future prediction can guide manipulation while
skipping it elsewhere. 

\section{Conclusion}

In this work, we presented Bind-WAM, a 
framework that uses language-object binding uncertainty to select
future imagination in WAMs. It achieves
strong performance on simulation and real-world tasks while
invoking joint inference at a minority of replanning steps.
These results support binding uncertainty as an effective signal
for selecting states that benefit from future
imagination in robot control.

Future work will explore incorporating task-outcome feedback into router
training to better identify when future imagination improves execution.
We also plan to 
extend the evaluation to a broader range of real-world
tasks and robot platforms to study the generality of uncertainty-guided
selective imagination. Exploring a broader range of interpretable signals beyond language-object binding uncertainty for selective future imagination is another promising direction.

\FloatBarrier
\bibliographystyle{IEEEtran}
\bibliography{references}

\end{document}